\documentclass[article,accept,moreauthors]{Definitions/mdpi}

\firstpage{1}
\pubvolume{1}
\issuenum{1}
\articlenumber{0}
\pubyear{2024}
\copyrightyear{2024}
\datereceived{ }
\daterevised{ } 
\dateaccepted{ }
\datepublished{\today}
\usepackage{subcaption,multirow,diagbox,fancyhdr,upgreek,siunitx}
\Title{An indoor DSO-based ceiling-vision odometry system for indoor industrial environments}
\TitleCitation{An indoor DSO-based ceiling-vision odometry system for indoor industrial environments}
\Author{Abdelhak Bougouffa $^{1,2,\ddagger,}*$\orcidA{}, Emmanuel Seignez $^{1}$, Samir Bouaziz $^{1}$ and Florian Gardes $^{2}$}
\AuthorNames{Abdelhak Bougouffa, Emmanuel Seignez, Samir Bouaziz and Florian Gardes}
\AuthorCitation{Bougouffa, A.; Seignez, E.; Bouaziz, S.; Gardes, F.}
\address{%
$^{1}$ \quad Université Paris-Saclay, ENS Paris-Saclay, CNRS, SATIE, 91190, Gif-sur-Yvette, France.; firstname.lastname@universite-paris-saclay.fr\\
$^{2}$ \quad ez-Wheel - IDEC Corp., 16400, La Couronne, France; f.gardes@ez-wheel.com}
\corres{Correspondence: abdelhak.bougouffa@gmail.com}
\firstnote{Current address: Astek Technology, 92100, Boulogne-Billancourt, France.}  
\conference{2022 17th International Conference on Control, Automation, Robotics and Vision (ICARCV) \citep{bougouffa-evaluation-2022}} 
\abstract{Autonomous Mobile Robots operating in indoor industrial environments require a
localization system that is reliable and robust. While Visual Odometry (VO) can
offer a reasonable estimation of the robot's state, traditional VO methods
encounter challenges when confronted with dynamic objects in the scene.
Alternatively, an upward-facing camera can be utilized to track the robot's
movement relative to the ceiling, which represents a static and consistent
space. We introduce in this paper Ceiling-DSO, a ceiling-vision system based on
Direct Sparse Odometry (DSO). Unlike other ceiling-vision systems, Ceiling-DSO
takes advantage of the versatile formulation of DSO, avoiding assumptions about
observable shapes or landmarks on the ceiling. This approach ensures the
method's applicability to various ceiling types. Since no publicly available
dataset for ceiling-vision exists, we created a custom dataset in a real-world
scenario and employed it to evaluate our approach. By adjusting DSO parameters,
we identified the optimal fit for online pose estimation, resulting in
acceptable error rates compared to ground truth. We provide in this paper a
qualitative and quantitative analysis of the obtained results.}
\date{}
\title{An indoor DSO-based ceiling-vision odometry system for indoor industrial environments}
\usepackage{natbib}
\begin{document}

\section{Introduction}
\label{sec:org0daa34f}
\label{introduction}

State estimation remains an open research problem in the field of mobile
robotics. The ability of a mobile robot to accomplish autonomous mobility tasks
depends on its knowledge of its position and orientation, referred to as its
pose, in space.

In indoor industrial environments, mobile robots are commonly classified into
two main categories: \emph{Automated Guided Vehicles (AGVs)} and \emph{Autonomous Mobile
Robots (AMRs)}. In one hand, AGVs rely on a dedicated infrastructure for
navigation. One approach is \emph{active guiding} \citep{qi-application-2015}, where,
for example, wires are embedded under the ground's surface to transmit wireless
signals that the robot detects and follows. Alternatively, \emph{passive guidance}
methods \citep{borenstein-omnimate-2000,han-system-2013} can be employed,
involving magnetic tapes, RFID tags, or visual lanes.

On the other hand, AMRs require little or no prior knowledge about the
environment. These robots rely on their sensors to perceive and comprehend the
surrounding environment, allowing them to estimate their ego-motion and
navigate. Over the past decade, there has been a significant interest in
developing AMRs. Their flexibility and ease of commissioning make them highly
suitable for various industrial applications.

In indoor robotics, several sensors can be used for estimating a robot's pose.
One commonly used method consists of utilizing wheel encoders to incrementally
estimate the robot's motion, which we call \emph{odometry}. However, the mechanical
coupling, the wheel slippage, and the absence of external correction make the
odometry unreliable for long-term usage \citep{siegwart-introduction-2011} due
to the drift caused by accumulating errors over time.

Alternatively, cameras can be employed to calculate the incremental changes in a
moving robot's position and orientation, a technique commonly referred to as
\emph{Visual Odometry (VO)} \citep{scaramuzza-visual-2011}. The term \emph{``visual odometry''}
was first introduced by Nister et al. \citep{nister-visual-2004}, it involves
incrementally estimating camera motion from a perceived stream of images.

From a conceptual standpoint, Visual Odometry approaches can be categorized into
two main families: \emph{indirect (feature-based) methods}
\citep{mur-artal-orbslam2-2017,awangsalleh-swift-2018,davison-monoslam-2007,awangsalleh-longitudinal-2019}
and \emph{direct methods} \citep{engel-direct-2018,forster-svo-2014}. Indirect methods
necessitate a preprocessing step that involves extracting a set of \emph{features}
using \emph{feature detection} techniques. These features are then used to minimize
\emph{geometric error}. On the other hand, direct methods use the \emph{raw pixel intensity
values} to minimize \emph{photometric error}, eliminating the need for explicit feature
extraction. Another classification criterion for VO approaches is the density of
their estimated 3D geometries. VO can estimate a \emph{dense} 3D geometry
\citep{geiger-stereoscan-2011}, a \emph{sparse} one
\citep{mur-artal-orbslam2-2017,davison-monoslam-2007}, or a \emph{semi-dense} geometry
\citep{engel-lsdslam-2014}.

\emph{Visual Odometry} (VO) and \emph{Visual Simultaneous Localization and Mapping (vSLAM)}
differ in their map management strategy. VO focuses only on maintaining
consistency in a partial local map and utilizes it to incrementally estimate
robot motion. In contrast, vSLAM algorithms aim to provide, alongside the
estimated poses, a globally consistent map, incorporating corrections on loop
closing \citep{scaramuzza-visual-2011}.

In the context of our mobile industrial robot, which operates in a highly
dynamic environment with multiple robots and humans sharing the same space, the
VO estimation can be challenging. Indeed, the presence of moving objects in the
scene necessitates their detection and exclusion while calculating the robot's
ego-motion \citep{sheng-dynamicdso-2020,kim-effective-2016}. This task tends to
be complicated and resource-intensive.

To address the challenges posed by indoor dynamic environments, a viable
solution involves using an upward-facing camera to monitor ceiling patterns and
track the movements of the robot. Wooyeon et al. were the first to propose such
an approach \citep{wooyeon-cvslam-2005}. They employed a monocular upward-facing
camera along with the Harris corner detector to extract corner features from the
captured images. These corners serve as landmarks within an SLAM framework based
on the \emph{Extended Kalman Filter (EKF)}. This method incorporates a multi-view
representation of the landmarks to improve data association and enhance the
accuracy of the system.

Ceiling vision-based odometry and SLAM approaches often rely on making
assumptions about the specific shapes and patterns observable on the ceiling.
Kim et al. \citep{kim-new-2013} have taken advantage of this by exploiting known
ceiling landmark classes (circular landmarks representing objects like \emph{lamps},
\emph{speakers}, \emph{fire alarms}, etc.). This valuable information is used in a FastSLAM
framework to effectively track the robot's pose and construct the map.

Another research effort by Hwang and Song \citep{hwang-monocular-2011} used a
similar approach, where their system detects three distinct types of landmarks:
\emph{corners}, \emph{circular lamps}, and \emph{doors}. Corner features are extracted using the
\emph{Features from Accelerated Segment Test (FAST)} detector. The lamps are identified
by locating the brightest patches in the ceiling image by using a \emph{Canny} edge
detector and performing geometric circle fitting on the bright zone. Doors are
recognized by detecting two vertical lines intersected by a horizontal line.
These detected features are then integrated into an EKF-SLAM framework, making
use of the robot's odometry during the prediction step.

Alternative methods involve making assumptions about the boundaries of the
ceiling. Choi et al. \citep{choi-efficient-2012,choi-efficient-2014} proposed
an approach using an EKF-SLAM framework with a monocular camera. They used the
boundaries between the ceiling and the walls, represented as lines with
intersection constraints, to construct a features map. The authors took
advantage of the fact that, in their test environment, ceiling regions dominate
most of the images captured from the ceiling-view camera, while walls occupy the
remaining portions.

Similarly, other approaches adopt the same assumption and focus on extracting
the ceiling boundaries, which are then compared to a known building blueprint
using a \emph{Monte Carlo Localization (MCL)} approach while employing an observation
model based on \emph{Ceiling Space Density (CSD)} \citep{ribacki-visionbased-2018}.

In other methods, distinct visual tags are carefully placed on the ceiling to
simplify the localization process. Li et al. \citep{li-improved-2018} designed
artificial markers that are easily detectable and placed them on the ceiling. By
processing the images captured from the ceiling camera, these markers are
detected, identified, and used to determine the camera's position and
orientation with respect to them. The obtained information is then incorporated
in a graph-based optimization algorithm, implemented within a Bayesian
estimation framework.

However, all the assumptions discussed above often do not hold in open
warehouses or industrial spaces. The vast working spaces may be too wide to
capture both the ceiling and the walls in a single image. Moreover, ceilings in
such environments tend to be high and may consist of inclined surfaces or be
distributed across different height levels. To address these challenges, we
propose employing a generic visual odometry approach using a ceiling-vision
camera. Specifically, we have chosen to make use of the \emph{Direct Sparse Odometry
(DSO)}. Being a direct approach, DSO allows us to track the camera's movement
even in regions with limited distinct features. Our objective is to propose and
evaluate a generic framework for ceiling observation that minimizes assumptions
about the observable shapes or landmarks on the ceiling.

In this paper, which extends our previous work
\citep{bougouffa-evaluation-2022}, we present and evaluate Ceiling-DSO, a
ceiling-vision odometry system based on \emph{Direct Sparse Odometry (DSO)}. Our
experiments were conducted in a real-world environment using a modular
industrial mobile robot platform. We systematically varied the DSO parameters
and carefully examined their impact on the system's performance and the quality
of the estimated trajectories. To evaluate the estimated trajectories, we
compared them with those obtained from a LiDAR-based SLAM solution, used as
ground truth. Furthermore, we offer both qualitative and quantitative analyses
of the obtained results.

The rest of this paper is structured as follows: Section \ref{sec:ceiling-dso}
introduces the original formulation of DSO and outlines the assumptions made in
implementing Ceiling-DSO. In section \ref{sec:experiment}, we provide details about
the experiment setup, including the robot's sensors, the dataset generated for
validation purposes, and the methodology employed in conducting the experiments.
The results of our study are presented and discussed in section \ref{sec:results}.
Finally, section \ref{sec:conclusion} summarizes the conclusions of this work and
discusses potential future directions.
\section{Ceiling Direct Sparse Odometry}
\label{sec:org8177c14}
\label{sec:ceiling-dso}

The \emph{Direct Sparse Odometry (DSO)}, introduced by Engel et al.
\citep{engel-direct-2018}, is a monocular visual odometry method that falls
under the category of direct methods. Unlike feature-based visual odometry
approaches that rely on a limited set of features for motion estimation, direct
methods use information from all pixels in the image. Direct methods can be
further classified into three categories based on the density of the generated
point cloud: \emph{dense}, \emph{semi-dense} and \emph{sparse}.

Dense methods leverage all pixels in the image to generate a high-density point
cloud. However, this typically leads to computationally intensive algorithms due
to the large amount of data being processed.

Semi-dense methods aim to reduce the number of processed by selecting a subset
of points for processing. These methods strike a balance between accuracy and
computational efficiency.

Sparse methods only consider a small subset of points for processing. This
approach further reduces the computational load but may sacrifice some level of
detail in the generated point cloud.

These different categories of direct methods offer varying trade-offs between
accuracy and computational complexity, allowing researchers to choose an
approach that best suits their specific requirements.

DSO belongs to the third category, it employs a gradient-based approach and
uniformly select candidate points from regions with high contrast. By
strategically choosing these points, DSO aims to enhance the estimation of
camera motion and achieve accurate odometry while keeping the amount of
processed data suitable for real-time applications. A \emph{photometric error
minimization} process is performed on these selected candidate points.

In the context of direct visual odometry/SLAM, the use of a \emph{global shutter}
camera is preferable for capturing frames simultaneously. This type of camera
ensures that all pixels of an image are acquired at the same instant. On the
other hand, \emph{rolling shutter} cameras capture image pixels sequentially, which can
introduce a systematic drift in the state estimation, making these type of
cameras unsuitable for DSO. However, this effect can be accounted for in the
optimization process by measuring and compensating it. For instance, Schubert et
al. \citep{schubert-direct-2018} introduced a DSO-based approach that
incorporates the rolling shutter constraint and estimates the capture time,
thereby improving trajectory estimation when using a rolling-shutter camera.

In this work, we adopt the same formulation as the original DSO
\citep{engel-direct-2018}. We model the camera sensor as a pinhole camera and
utilize the intrinsic matrix \(\mathbf{K}\) for geometric projection. This
projection maps 3D points onto the image plane \(\Omega\) and is denoted as
\(\Pi_{\mathbf{K}}\). Additionally, we use the back-projection function
\(\Pi_{\mathbf{K}}^{-1}\) to convert a 2D point in the image plane along with its
depth information into the corresponding 3D world coordinates.

\begin{equation*}
\label{eq:dso-matrix-and-proj-params}
\text{with:} \quad
\begin{cases}
    \mathbf{K} \in \mathbb{R}^{3 \times 3} & \quad \text{Cameras intrensic matrix} \\
    \Omega \in \mathbb{R}^{2 \times 2} & \quad \text{Image plane} \\
    \Pi_{\mathbf{K}} : \mathbb{R}^{3} \rightarrow \Omega & \quad \text{Projection onto the image plane $\Omega$} \\
    \Pi_{\mathbf{K}}^{-1} : \Omega \rightarrow \mathbb{R}^{3} & \quad \text{Back-projection from $\Omega$ to the 3D world}
\end{cases}
\end{equation*}

In line with DSO, we consider \emph{photometric calibration}
\citep{engel-photometrically-2016}. For each frame ``\(i\)'', the camera observes
the raw intensity \(I_{i}^{RAW}\) of a pixel \(\boldsymbol{x}\). This intensity is
defined as a function of various factors, including irradiance \(B_{i}\), exposure
time \(t_{i}\), a non-linear response function \(G : \mathbb{R} \rightarrow [0, 255]\), and
lens attenuation (vignetting) \(V : \Omega \rightarrow [0, 1]\).

\begin{equation}
\label{eq:dso-intensity-combined}
I_{i}^{RAW} ( \boldsymbol{x} ) = G( t_{i} V( \boldsymbol{x} ) B( \boldsymbol{x} ) )
\end{equation}

\begin{equation*}
\label{eq:dso-photo-calib-params}
\text{with:} \quad
\begin{cases}
    i \in \mathbb{N} &  \quad \text{A number to index frames} \\
    \boldsymbol{x} \in \Omega & \quad \text{A pixel on the image plane $\Omega$} \\
    t_{i} \in \mathbb{R} & \quad \text{Exposure time of the frame $i$} \\
    I_{i}^{RAW} : \Omega \rightarrow [0, 255] & \quad \text{Raw pixel intensities image} \\
    I_{i} : \Omega \rightarrow [0, 255] & \quad \text{Photometrically corrected image} \\
    G : \mathbb{R} \rightarrow [0, 255] & \quad \text{Non-linear response function} \\
    V : \Omega \rightarrow [0, 1] & \quad \text{Lens attenuation} \\
    B_{i} : \Omega \rightarrow [0, 1] & \quad \text{Irradiance}
\end{cases}
\end{equation*}

By reverting the non-linear function and compensating for the vignetting effect
on the image, we can calculate the photometrically corrected image \(I_{i}\) as
shown in equation (\ref{eq:dso-intensity}), which is derived from equation
(\ref{eq:dso-intensity-combined}).

\begin{equation}
\label{eq:dso-intensity}
I_{i}( \boldsymbol{x} ) \triangleq
t_{i }B_{i}( \boldsymbol{x} ) =
\frac{G^{-1} ( I_{i}^{RAW} ( \boldsymbol{x} ) )}{V ( \boldsymbol{x} ) }
\end{equation}

The overall photometric error across all frames can be expressed as follows:

\begin{equation}
\label{eq:full-photometric-error}
E_{\text{photometric}} = \sum_{i \in N} \sum_{\mathbf{p} \in P_{i}} \sum_{j \in obs(\mathbf{p})} E_{i,\mathbf{p},j}
\end{equation}

Considering a total of \(N\) frames, let \(P_{i}\) denote the set of points observed
in the \(i\text{-th}\) frame. The variable \(j\) iterates over \(obs(\mathbf{p})\),
representing all frames in which the point \(\mathbf{p}\) is visible. The partial
photometric error term \(E_{i,\mathbf{p},j}\) is defined as a weighted sum of
Huber \citep{huber-robust-1964} norms, calculated within a neighborhood pattern
of points surrounding the point \(\mathbf{p}\), denoted by set
\(\mathcal{N}_{\mathbf{p}}\).

\begin{equation}
\label{eq:partial-photometric-error}
E_{i,\mathbf{p},j} = \sum_{\mathbf{p} \in \mathcal{N}_{\mathbf{p}}} w_{\mathbf{p}} \bigg\lVert (I_{j}[\mathbf{p}'] - b_{j})
- \frac{t_{j} e^{a_{j}}}{t_{i} e^{a_{i}}}(I_{i}[\mathbf{p}] - b_{i}) \bigg\rVert_{\gamma}
\end{equation}

\begin{align*}
\label{eq:huber-loss}
\text{with Huber's norm~} &
\lVert \alpha \rVert_{\gamma}
\triangleq
\begin{cases}
\frac{1}{2} \alpha^{2} & \text{for~} | \alpha | < \gamma \\
\gamma \cdot (| \alpha | - \frac{1}{2} \gamma) & \text{otherwise}
\end{cases}
\\
\text{and~} &
w_{\mathbf{p}} \triangleq \frac{c^{2}}{c^{2} + \lVert \nabla I_{i} ( \mathbf{p} ) \rVert^{2}_{2}}
\text{~~with~} c \in \mathbb{R}
\end{align*}

The Huber norm is a hybrid norm, combining the \(\ell_{1}\) and \(\ell_{2}\) norms and
offering robustness against outliers while being differentiable throughout. As a
result, it serves as a well-suited error measurement for gradient-based
optimization.

The residual term within the summation (\ref{eq:partial-photometric-error}) accounts
for the discrepancy between the intensity values of point \(\mathbf{p}\) in the
current frame (\(i\)) and its intensity across all frames where it is observable
\(j \in obs(\mathbf{p})\). To account for the unknown exposure times, the
intensity is modeled as an affine brightness transfer function
\citep{engel-photometrically-2016}. The \(\mathbf{p}'\) in equation
(\ref{eq:partial-photometric-error}) represents the projection of point \(\mathbf{p}\),
which is seen in the frame (\(i\)), onto the frame (\(j\)) with an estimated depth
of \(d_\mathbf{p}\). The projection matrix relies on the partial camera motion
transform \(\Delta \mathbf{T}_{j,i}\), which describes the transformation between
camera poses \(\mathbf{T}_{i}\) and \(\mathbf{T}_{j}\).

\begin{equation}
\label{eq:dso-point-projection}
\mathbf{p}' = \Pi_{\mathbf{K}}(\mathbf{R}
\Pi_{\mathbf{K}}^{-1}(\mathbf{p}, d_{\mathbf{p}}) + \mathbf{t})
\end{equation}

\begin{equation*}
\label{eq:dso-delta-transform}
\text{with:} \quad
\Delta \mathbf{T}_{j,i}
=
\begin{bmatrix}
\mathbf{R} & \mathbf{t} \\ 0 & 1
\end{bmatrix}
= \mathbf{T}_{j} \mathbf{T}_{i}^{-1}
\end{equation*}

The minimization of the error term of equation (\ref{eq:partial-photometric-error}) is
achieved using a Gauss-Newton optimization algorithm, which iterates through 6
steps. This optimization is performed on the Lie algebra \(\mathfrak{se}(3)\),
where the \(\text{left-}\oplus\) operator is defined as follows:

\begin{equation}
\label{eq:dso-opt-left-plus}
\oplus : \mathfrak{se}(3) \times \text{SE}(3) \rightarrow \text{SE}(3)
\end{equation}

\begin{equation*}
\text{with:} \quad
\begin{cases}
\mathbf{x}_{i} \in \mathfrak{se}(3) , \hspace{3mm} \mathbf{T}_{i} \in \text{SE}(3) \\
\mathbf{x}_{i} \oplus \mathbf{T}_{i} \triangleq e^{\widehat{\mathbf{x}_{i}}} \cdot \mathbf{T}_{i}
\end{cases}
\end{equation*}

For more details about Lie groups in the context of robotics and state
estimation, readers are encouraged to refer to the paper by Solà et al.
\citep{sola-micro-2020}. This work delves into the applications of the Lie
groups theory in robotics and state estimation, providing valuable insights for
researchers in the field.

In the optimization process, the parameters to be optimized are denoted as \(\zeta \in
\text{SE}(3)^{n} \times \mathbb{R}^{m}\). These parameters encompass both geometric
aspects, such as poses, inverse depth values, and camera intrinsics, as well as
photometric parameters, represented by the affine brightness parameters \((a_{i},
b_{i})\). Within the rigid motion manifold \(\text{SE}(3)\), we have an initial
evaluation point denoted as \(\zeta_{0}\), and the accumulated delta updates are
represented by \(\mathbf{x} \in \mathfrak{se}(3)^{n} \times \mathbb{R}^{m}\). The current
state estimate can be obtained by \(\zeta = \mathbf{x} \oplus \zeta_{0}\), using the
\(\text{left-}\oplus\) operator, which is extended beyond \(\text{SE}(3)\) elements as a
regular addition.

This optimization process of the photometric error (\ref{eq:full-photometric-error})
is formulated as a Gauss-Newton system, which is defined as follows:

\begin{align}
\label{eq:dso-opt-sys}
\mathbf{H} &= \mathbf{J}^{T} \mathbf{W} \mathbf{J} \\
\mathbf{b} &= -\mathbf{J}^{T} \mathbf{W} \mathbf{r}
\end{align}

In the equations above, the vector \(\mathbf{r} \in \mathbb{R}^{n}\) represents the
collection of residuals, \(\mathbf{W} \in \mathbb{R}^{n \times n}\) is a square diagonal
matrix with each diagonal element representing a weight factor, and \(\mathbf{J}
\in \mathbb{R}^{n \times d}\) is the Jacobian matrix of the residuals vector
\(\mathbf{r}\).

Consider \(r_{k}\) as an individual residual from the vector \(\mathbf{r}\),
and let \(J_{k}\) represent its corresponding row in the Jacobian matrix
\(\mathbf{J}\). The residual \(r_k\) contains: the camera
poses at frames ``\(i\)'' and ``\(j\)'', denoted respectively, \(\mathbf{T}_{i}\) and \(\mathbf{T}_{j}\); the inverse depth \(d_{\mathbf{p}}\), the camera intrinsics \(\mathbf{K}\), and
the affine brightness parameters \(a_{i}\), \(a_{j}\), \(b_{i}\) and \(b_{j}\).

\begin{equation}
\label{eq:resid}
\text{with:} \quad
(\mathbf{T}_{i} , \mathbf{T}_{j} , d_{\mathbf{p}} , a_{i} , a_{j} , b_{i} , b_{j}) = \mathbf{x} \oplus \zeta_{0}
\end{equation}

\begin{align}
\label{eq:dso-opt-residual}
r_{k} &= ( I_{j} [ \mathbf{p}' ( \mathbf{T}_{i} , \mathbf{T}_{j} , d , \mathbf{K} ) ] - b_{j} )
- \frac{ t_{j} e^{a_{j}} }{ t_{i} e^{a_{i}} } ( I_{i} - b_{i} ) \\
\mathbf{J}_{k} &= \frac{ \partial r_{k} ( ( \updelta + \mathbf{x} ) \oplus \zeta_{0} ) }{ \partial \updelta }
\end{align}

During the Gauss-Newton optimization process in DSO, a sliding window of \(N_{f}\)
keyframes is employed. Any points that fall outside this window are
marginalized, meaning they are no longer actively considered in the
optimization, which helps keeping the execution time bounded.

In our Ceiling-DSO implementation, we simplified the DSO formulation by assuming
a linear response function \(G(\boldsymbol{x})\).
Additionally, we used lenses without vignetting, so:

\begin{equation}
\label{eq:ceiling-dso-assumptions}
\begin{cases}
    \forall \boldsymbol{x} \in \Omega : G(\boldsymbol{x}) = \boldsymbol{x} \\
    \forall \boldsymbol{x} \in \Omega : V(\boldsymbol{x}) = 1
\end{cases}
\end{equation}
\section{Experiment}
\label{sec:org6bda87c}
\label{sec:experiment}
\subsection{Experimental platform}
\label{sec:org99852ae}
\begin{figure}[htbp]
\centering
\includegraphics[width=0.75\textwidth]{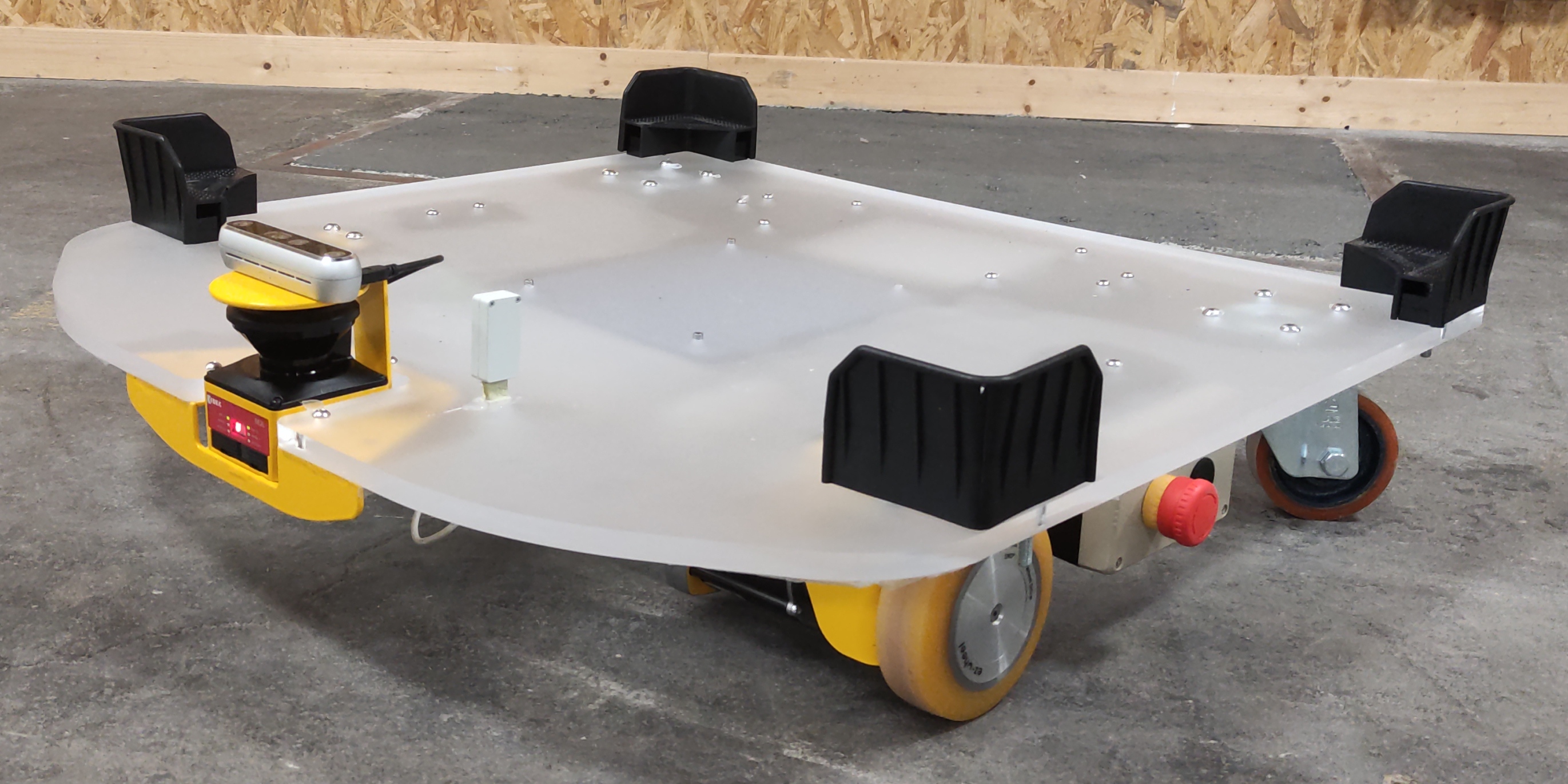}
\caption{\label{fig:platform}The SWD Starter Kit}
\end{figure}

In our experiments, we employed a mobile industrial robot prototype (figure
\ref{fig:platform}). This platform is a modular, differential-drive robot that
utilizes two self-contained motorized wheels powered by the \emph{ez-Wheel Safety
Wheel Drive (SWD\texttrademark{})} technology. This prototype represents an evolution of our
previously validated SmartTrolley platform \citep{bougouffa-smarttrolley-2020}.
Its primary objective is to facilitate the movement of heavy loads indoors,
particularly in industrial environments. We carefully designed and sized the
robot to handle a maximum load capacity of 2 tonnes (\(2000 \si{kg}\)), however,
in this experiment, we used a plexiglass structure to facilitate moving the
robot during the test phase. To prioritize safety, the robot is programmed to
operate solely at low speeds of approximately \(5 \si{km \cdot h^{-1}}\) (\(1.4 \si{m \cdot
s^{-1}}\)).

The robot is equipped with a pair of incremental wheels encoders, two cameras,
the first is an \emph{Intel\textregistered{} RealSense\texttrademark{} D435i} facing
forward, while the second is an \emph{Intel\textregistered{} RealSense\texttrademark{}
455} facing upward. Furthermore, the robot integrates an \emph{IDEC S2L} safety LiDAR,
covering a maximal range of \(30 \si{m}\).

The platform is equipped with a powerful embedded industrial computer,
specifically the \emph{Neousys Nuvo-7002LP}, featuring an 8th Generation
\emph{Intel\textregistered{} Coffee lake Core\texttrademark{} i5} processor and 16GB
DDR4 2666/2400 SDRAM. The computer operates on the \emph{Ubuntu 20.04} operating
system, complemented by the robotics middleware \emph{ROS Noetic}. This used this
embedded computer to run the differential-drive kinematic model for robot
control and for gathering raw sensor data. To enable teleoperation, we equipped
the computer with a wireless transceiver and used a wireless joystick to control
the robot remotely.

The platform leverages the safety LiDAR's low-level obstacle detection
capabilities. We established two distinct detection zones for enhanced safety
measures. The first zone, referred to as the \emph{Safety-Limited Speed (SLS)} zone, is
activated when the robot approaches an obstacle, allowing for reduced speed. The
second zone, known as the \emph{Safe Direction Indication (SDI)} zone, is smaller and
prevents the robot from moving towards nearby obstacles. These detection and
response behaviors have been implemented at the lowest level of the system (in
the wheels' microcontrollers). To facilitate robust and secure integration, the
LiDAR transmits the SDI and SLS signals directly to the motorized wheels via
secure \emph{Output Signal Switching Device (OSSD)} outputs.
\subsection{Dataset}
\label{sec:org64cbacb}
Our experiment took place within a spacious indoor open area measuring \(21 \times 15
\si{m}\). Throughout the experiment, we gathered raw sensor data from various
sources, including odometry readings, stereo up-facing camera images,
forward-facing camera images, and LiDAR range measurements.

\begin{figure}[htbp]
\centering
\includegraphics[width=0.75\textwidth]{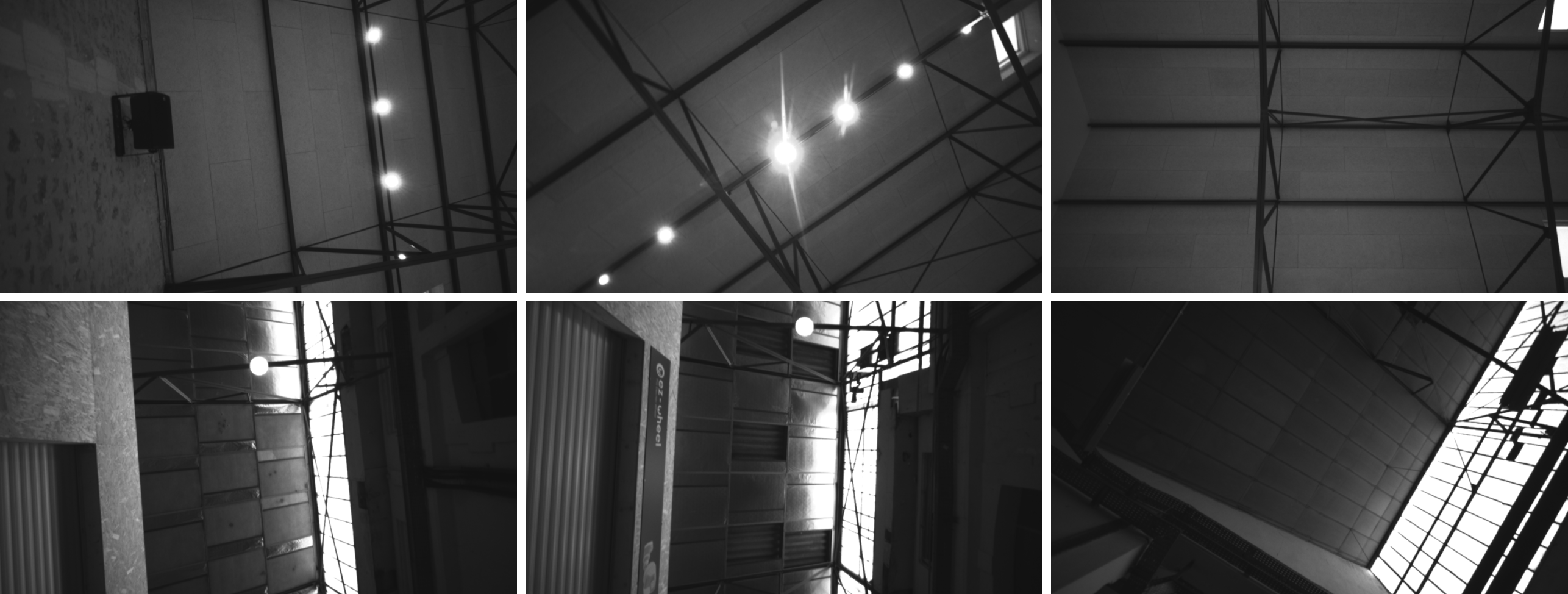}
\caption{\label{fig:ceiling}Sample images obtained from the up-facing camera offer a visual appearance of the test environment's ceiling.}
\end{figure}

In our test environment, the ceiling in inclined (triangular structure), with
varying heights between 4 and 6 meters. The ceiling's characteristics, such as
shapes and landmarks, may differ across different regions of the test area. For
reference, figure \ref{fig:ceiling} provides example images captured by the up-facing
camera, showcasing the visual appearance of the ceiling in our experimental
setup.

\subsection{Methodology}
\label{sec:orgf6aedf3}
To assess the performance of the Ceiling-DSO, we conducted an evaluation using a
series of sequences from our collected dataset. Our analysis focused on
investigating the impact of different factors, namely the \emph{size of input images},
the \emph{frame rate}, and \emph{the maximum size of the optimization window}. We aimed to
determine how these parameters affect both the execution time and the quality of
the estimated trajectory. By identifying favorable parameter combinations
through this evaluation, our objective was to optimize the Ceiling-DSO for
real-time applications.

\begin{figure}[htbp]
\centering
\includegraphics[width=0.75\textwidth]{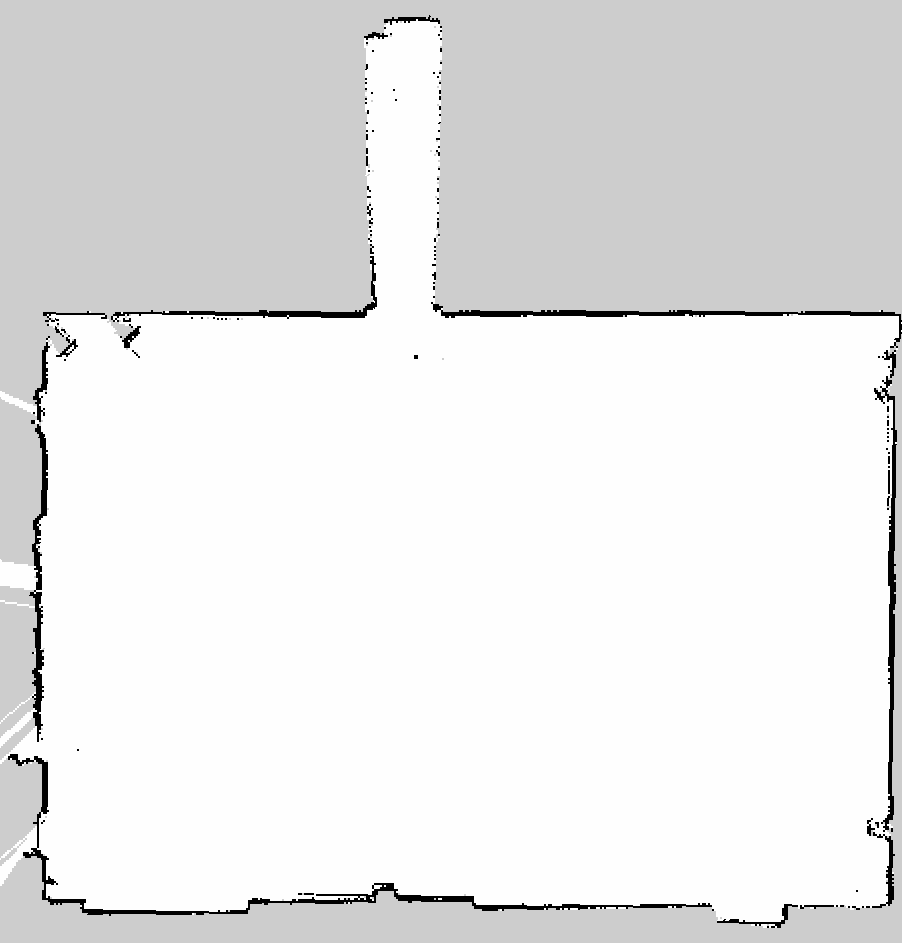}
\caption{\label{fig:gt-map}The test environment's map}
\end{figure}

Utilizing the data obtained from the LiDAR, we computed a ground truth
trajectory by employing the open-source LaMa SLAM algorithm
\citep{pedrosa-fast-2020}. The resulting map of the test environment, generated
through this SLAM algorithm, is visually depicted in figure \ref{fig:gt-map}.

We carried out our experiment on a planar surface, utilizing a camera and a 2D
LiDAR that were fixed on the robot. We refer to the ground truth 2D trajectory
in 2D as \(\mathcal{G}\). As DSO utilizes a monocular camera, its estimated
trajectory, denoted as \(\mathcal{P}\), is only valid up to a scale factor \(\lambda \in
\mathbb{R^{+}}\). While the DSO trajectory is 3-dimensional, the ground truth
trajectory is limited to 2D. Given that the robot's movement occurs only on a
planar surface, we can effectively align and compare the two trajectories.

In order to align the estimated trajectories for subsequent comparison, we adopt
a method similar to the approach proposed by Zhang and Scaramuzza
\citep{zhang-tutorial-2018}. Initially, we synchronize the two trajectories
using the global clock provided by ROS. Denoting \(\mathcal{G}'\) and
\(\mathcal{P}'\), the \(n\) ground truth and visual odometry synchronized positions,
respectively, defined as follows:

\begin{equation}
\label{eq:taj-sync}
\begin{aligned}
\mathcal{G}' &= \{ \mathbf{g} \; \vert \; \lVert t_{\mathbf{p}} - t_{\mathbf{g}} \rVert < \tau \} \\
\mathcal{P}' &= \{ \mathbf{p} \; \vert \; \lVert t_{\mathbf{p}} - t_{\mathbf{g}} \rVert < \tau \}
\end{aligned}
\text{\hspace{5mm}with:~}
\begin{cases}
\forall \mathbf{g} \in \mathcal{G} \\
\forall \mathbf{p} \in \mathcal{P} \\
\tau \in \mathbb{R^{+}}
\end{cases}
\end{equation}

Here, \(t_{\mathbf{p}}\) and \(t_{\mathbf{g}}\) represent the timestamps of points
\(\mathbf{p}\) and \(\mathbf{g}\), respectively. The synchronization threshold is
denoted as \(\tau\).

We perform an estimation of the global similarity transformation denoted as
\(\mathbf{S}\) between the synchronized trajectories of the ground truth
(\(\mathcal{G}'\)) and visual odometry (\(\mathcal{P}'\)). This transformation is
expressed as \(\mathbf{S} = (\mathbf{R}, \mathbf{t}, \lambda)\), where \(\mathbf{R} \in
\text{SO}(3)\) represents the 3D rotation, \(\mathbf{t} \in \mathbb{R}^{3}\) denotes
the translation, and \(\lambda \in \mathbb{R^{+}}\) represents the scale.

The alignment process can be formulated as a least-squares problem, aiming to
minimize the discrepancy between the synchronized position pairs \(\mathbf{p}_{i}
\in \mathcal{P}'\) and \(\mathbf{g}_{i} \in \mathcal{G}'\). The optimal transformation
for global alignment, denoted as \(\mathbf{S}^{\star} = (\mathbf{R}^{\star},
\mathbf{t}^{\star}, \lambda^{\star})\), can be expressed as follows:

\begin{equation}
\label{eq:taj-argmin}
\mathbf{S}^{\star} = \underset{\mathbf{R},\mathbf{t},\lambda}{\arg\min} \sum_{i = 0}^{n}\lVert \mathbf{g}_{i} - ( \lambda \mathbf{R} \mathbf{p}_{i} + \mathbf{t} ) \rVert^{2}
\end{equation}

The estimated visual odometry (VO) trajectory, denoted as \(\mathcal{P}\), is
subsequently aligned with the optimal transformation \(\mathbf{S}^{\star}\). This
alignment process yields the scaled trajectory \(\mathcal{P}^{\star}\).

\begin{equation}
\label{eq:taj-scaled}
\forall \mathbf{p} \in \mathcal{P} : \mathcal{P}^{\star} = \{ \lambda^{\star} \mathbf{R}^{\star} \mathbf{p} + \mathbf{t}^{\star} \}
\end{equation}

To assess the performance, we utilize relative errors (REs) as metrics
\citep{zhang-tutorial-2018}. These REs quantify the disparities between the
ground truth and the trajectory generated by Ceiling-DSO. We then calculate the
Euclidean norm of the position error to provide an evaluation of the obtained
results.
\section{Results and discussion}
\label{sec:org3761cf6}
\label{sec:results}

We tested Ceiling-DSO on various sequences extracted from our dataset. For this
paper, we have chosen to showcase two specific sequences. The first sequence
represents a simple trajectory resembling a square shape, while the second one
demonstrates a loop closing trajectory featuring multiple smaller loops.

Initially, we present qualitative results by aligning and visually comparing the
aligned trajectories with the corresponding ground truth. Throughout this
analysis, we conducted a systematic evaluation by testing a total of 24
trajectories per sequence. These trajectories were generated by iterating over
all possible combinations of tested parameters, including two image sizes (\(848
\times 480\) or \(424 \times 240\)), four frame rates (3, 6, 15, or 30fps), and three maximum
optimization window sizes (5, 7, or 15). Such testing allows for a systematic
examination of the performance of the system across various parameter
configurations.

\begin{figure}[htbp]
\centering
\includegraphics[width=0.8\textwidth]{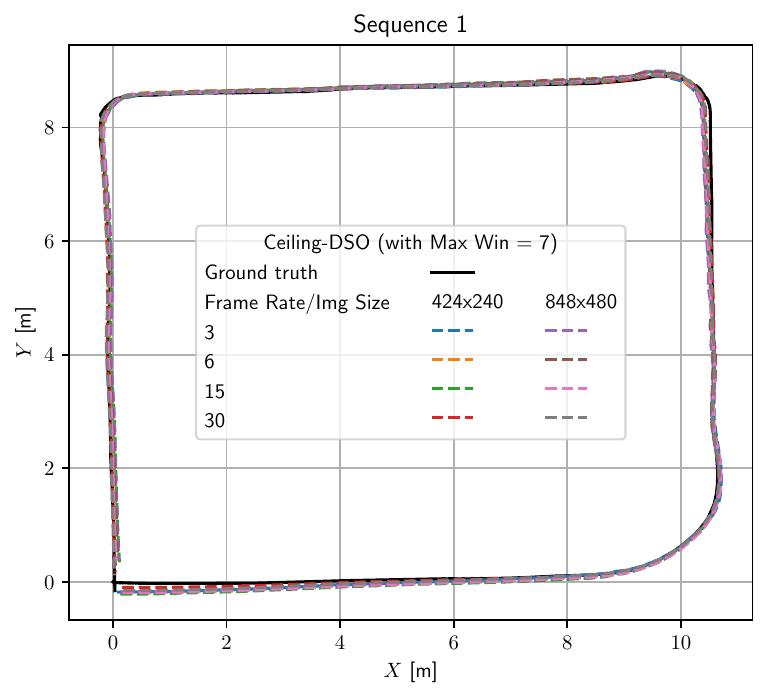}
\caption{\label{fig:traj-fps-seq1}Trajectories for various image sizes and frame rates at a fixed maximum window size of 7 \emph{(sequence 1)}.}
\end{figure}

\begin{figure}[htbp]
\centering
\includegraphics[width=0.8\textwidth]{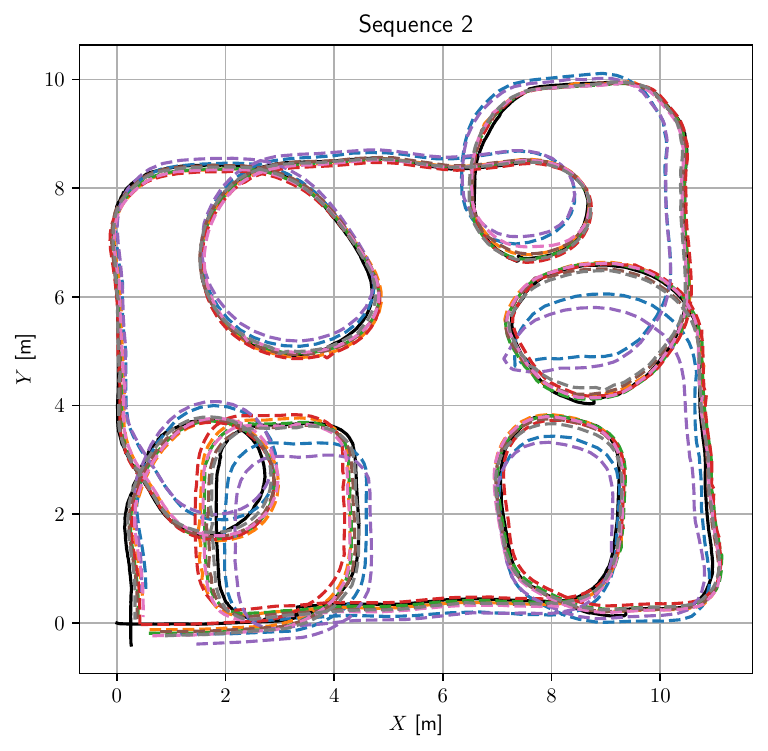}
\caption{\label{fig:traj-fps-seq2}Trajectories for various image sizes and frame rates at a fixed maximum window size of 7 \emph{(sequence 2)}.}
\end{figure}

When considering the estimated trajectories from the two sequences, figures
\ref{fig:traj-fps-seq1} and \ref{fig:traj-fps-seq2} showcase the impact of varying the frame
rate and image size while keeping the maximum optimization window size fixed
at 7. It is noted that reducing the frame rate below 15fps leads to an increase
in trajectory error, which becomes more pronounced in the case of the complex
trajectory of the second sequence in figure \ref{fig:traj-fps-seq2}. It is important to
highlight that the choice of frame rate is closely linked to the speed of the
robot. For robots operating at higher speeds, it becomes necessary to utilize
streams with higher frame rates. As DSO implements a coarse-to-fine matching
approach, increasing the image size can aid in refining the estimation. However,
our tests revealed no significant impact on the estimation accuracy when
reducing the image size from \(848 \times 480\) to \(424 \times 240\).

\begin{figure}[htbp]
\centering
\includegraphics[width=0.8\textwidth]{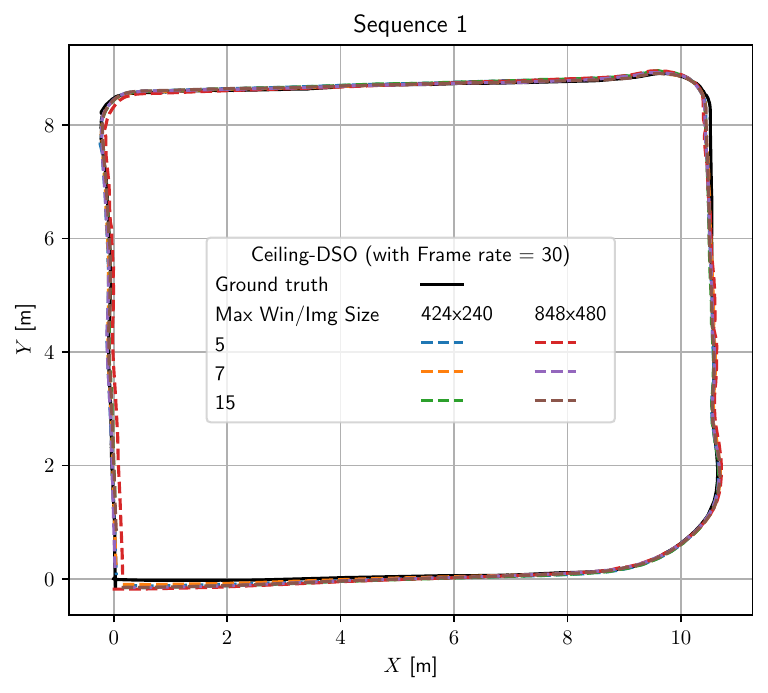}
\caption{\label{fig:traj-max-win-seq1}Trajectories for various image sizes and maximum window sizes at a fixed frame rate of 30 \emph{(sequence 1)}.}
\end{figure}

\begin{figure}[htbp]
\centering
\includegraphics[width=0.8\textwidth]{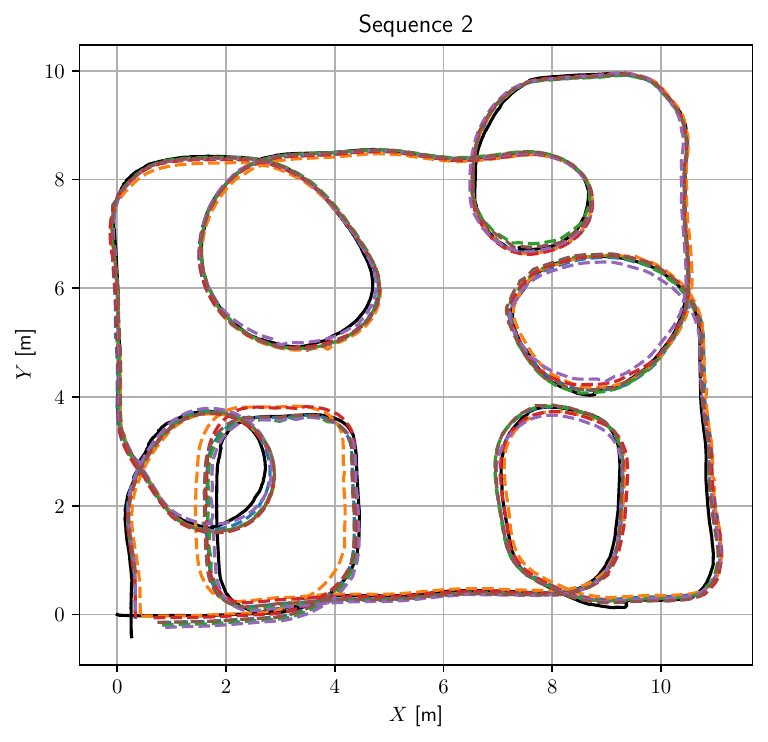}
\caption{\label{fig:traj-max-win-seq2}Trajectories for various image sizes and maximum window sizes at a fixed frame rate of 30 \emph{(sequence 2)}.}
\end{figure}

The impact of changing the image size and the maximum optimization window size
(while fixing the frame rate at 30fps) on the trajectories can be observed in
figures \ref{fig:traj-max-win-seq1} and \ref{fig:traj-max-win-seq2}. The trajectories
exhibit slight improvements when increasing the maximum window size. This can be
attributed to the local bundle adjustment performed during the optimization
step. Using a larger window size allows for improved accuracy, particularly in
more complex trajectories. However, our tests demonstrated that an acceptable
level of accuracy could be achieved with a maximum window size of 7.

\begin{figure}[htbp]
\centering
\includegraphics[width=0.7\textwidth]{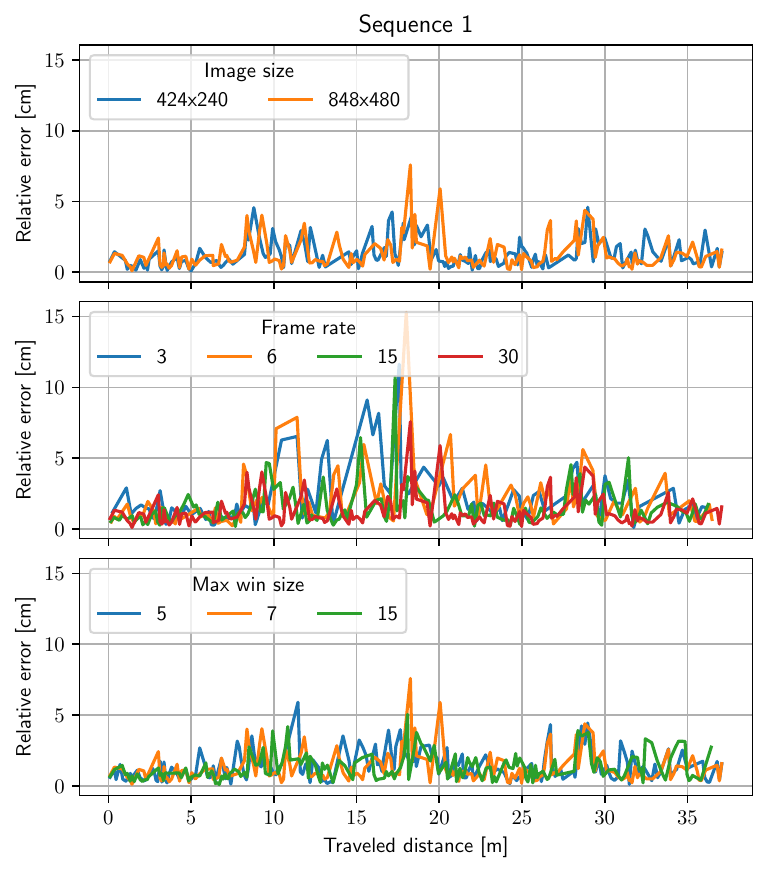}
\caption{\label{fig:err-rel-seq1}The relative error of the DSO trajectory compared to the ground truth for various frame rates, image sizes, and maximum optimization window sizes (Seq1).}
\end{figure}

\begin{figure}[htbp]
\centering
\includegraphics[width=0.7\textwidth]{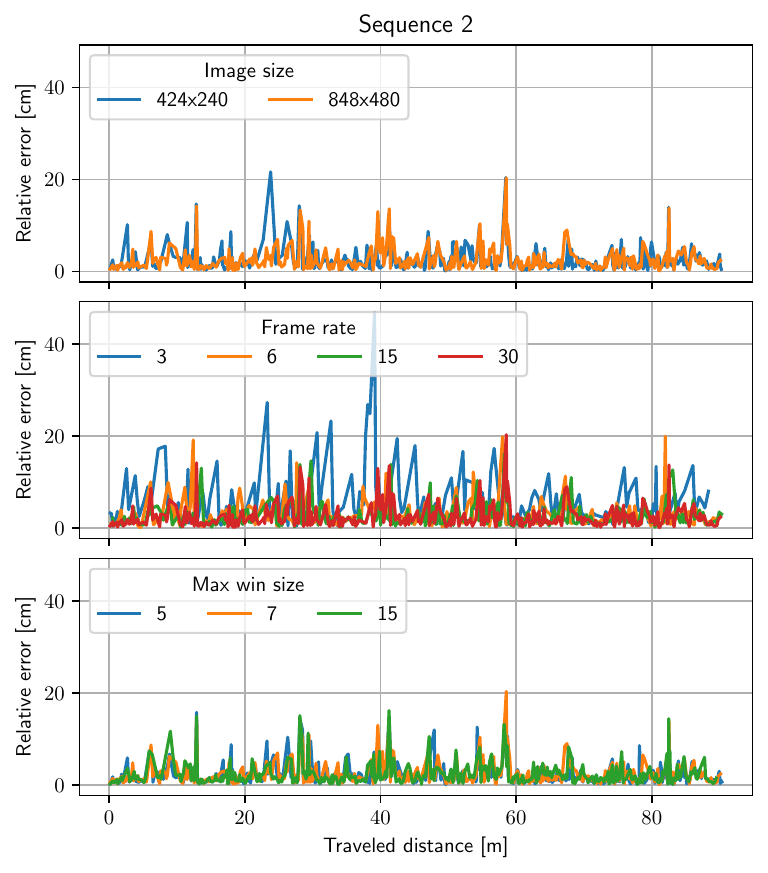}
\caption{\label{fig:err-rel-seq2}The relative error of the DSO trajectory compared to the ground truth for various frame rates, image sizes, and maximum optimization window sizes (Seq2).}
\end{figure}

For the quantitative analysis, we present the plot of the Euclidean norm of the
relative position error \(\epsilon_{i} = \sqrt{\Delta x^{2}_{i}+ \Delta y^{2}_{i} + \Delta z^{2}_{i}}\)
as a function of the traveled distance. Figures \ref{fig:err-rel-seq1} and
\ref{fig:err-rel-seq2} provide a summary of the relative error observed in the two
sequences, with each figure representing a variation of one parameter while
keeping the other two at their default values (defaults are, image size: \(848 \times
420\), frame rate: 30fps, and maximum window size: 7). It is noted from the
results that the most significant errors occurred when the frame rate was
decreased below 15fps. However, we did not observe any significant influence on
the relative error when varying the other parameters.

\begin{figure}[htbp]
\centering
\includegraphics[width=0.8\textwidth]{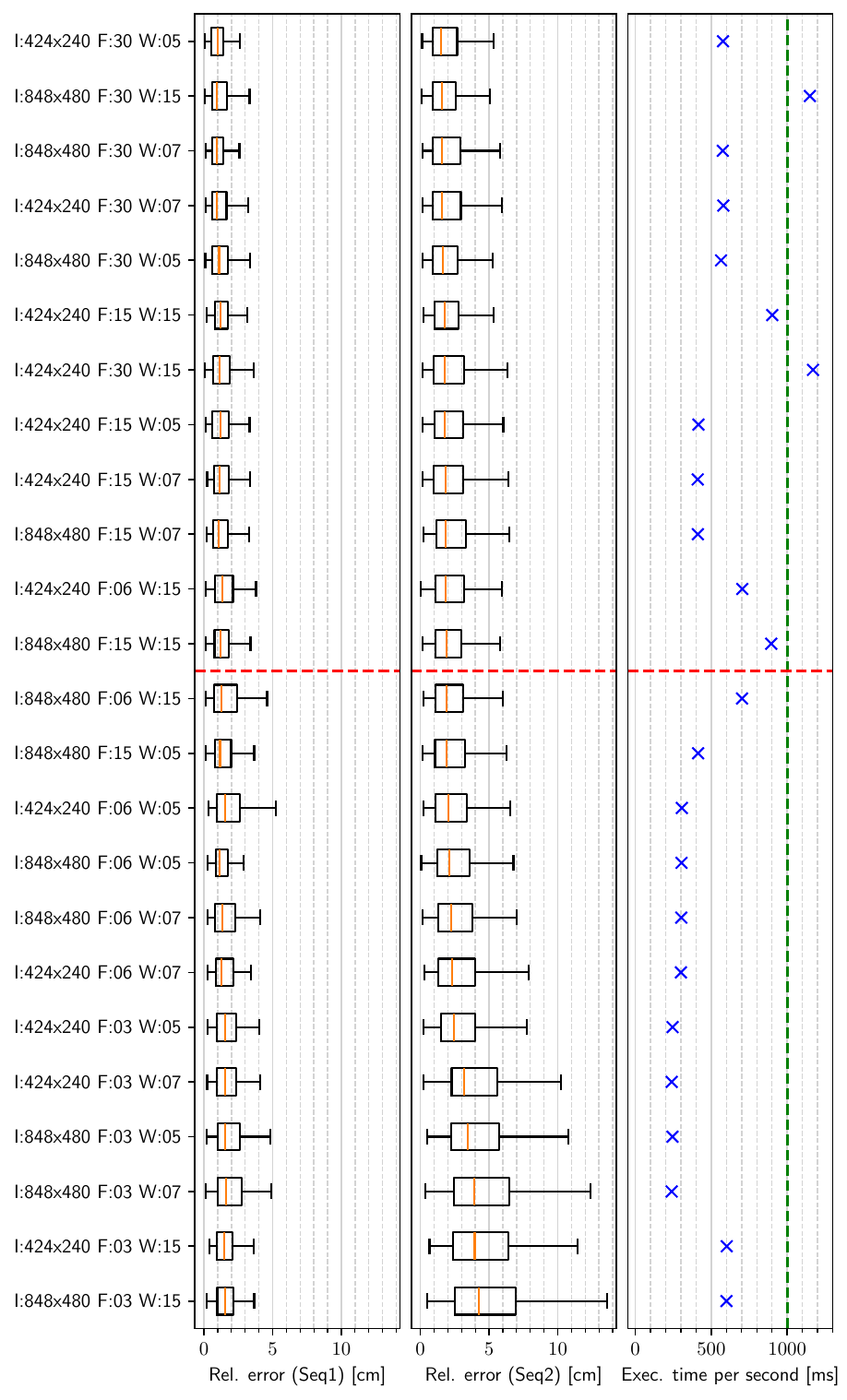}
\caption{\label{fig:err-rel-stats}Box plots of relative errors and calculation time for each parameter combination (I: \emph{image size}, F: \emph{frame rate}, W: \emph{optimization window size}). The boxes represent the first and third quartiles, with the orange line indicating the median. Whiskers depict the variability within \(1.5 \times \text{IQR}\). Boxes are sorted by Seq2's median relative errors.}
\end{figure}

In order to understand the distribution of relative errors across the entire
trajectory and for each parameter combination, we present in figure
\ref{fig:err-rel-stats}, which provides box plots that depict the relative error
distribution, highlighting key statistical measures for each combination of
tested parameters. The box represents the \emph{interquartile range (IQR)}, with the
first quartile \(Q_1\) and third quartile \(Q_3\) defining its boundaries. The
median, denoted as \(Q_2\), is represented by a line within the box. The whiskers
extend from the box to indicate the variability beyond the lower and upper
quartiles, within a range of \(1.5 \times \text{IQR} = 1.5 \times (Q_{3} - Q_{1})\). This
allows for a visual comparison of the relative error distributions for the two
sequences.

The right subfigure of the plot \ref{fig:err-rel-stats} provides the average execution
time per second, which is calculated as the product of the per-frame execution
time (\(t_F\)) in milliseconds and the frame rate (\(f\)) in hertz. These values are
obtained from table \ref{tab:execution-time}. The vertical green line positioned at
\(t=1000\si{ms}\) represents the real-time limit and serves as a reference for
determining whether the processing time remains within real-time constraints.

In figure \ref{fig:err-rel-stats}, a red dashed line is depicted to bisect the sorted
medians of the relative errors. We focus on the parameter combinations in the
first half, as they exhibit lower relative errors. Conversely, the second half,
which comprises combinations associated with higher relative errors, is mostly
characterized by frame rates below 15fps. Notably, the second trajectory
demonstrates more pronounced errors. Based on these observations, we can propose
a frame rate regulation strategy that aligns with the robot's movement.
Specifically, when the robot is moving along a straight line, a lower frame rate
can be employed, while an increased frame rate would be preferable during
steering maneuvers.

The average execution time per frame was measured for each combination of the
tested parameters. The measured timings (in milliseconds) in table
\ref{tab:execution-time} provide insights into the \emph{per-frame average execution time}.
Surprisingly, reducing the image size did not have a significant impact on the
execution time. However, increasing the optimization window size resulted in
longer execution times due to the increased workload in each optimization step.

Interestingly, decreasing the frame rate was associated with an increase in the
execution time per frame. This counter-intuitive observation can be explained by
the fact that the system creates keyframes (which are used in the optimization
process) more frequently in low frame rate settings. To support this
observation, table \ref{tab:keyframe-per-frame} presents the ratio of keyframes to
frames, revealing no correlation between the window size or image size and the
keyframe-to-frame ratio.

\begin{table}[!htbp]
\begin{minipage}{.48\textwidth}
\centering
\begin{tabular}{|r|c|c c c c|}
\hline
\multicolumn{1}{|c|}{\bf{Win.}} & \diagbox{\bf{Image}}{\bf{Fps}}
                                             & 3       & 6       & 15      & 30      \\ \hline
\multirow{2}{*}{5}
                             & $848 \times 480$   & $81.7$  & $50.6$  & $27.5$ & $18.8$ \\ \cline{2-6}
                             & $424 \times 240$   & $81.8$   & $51.1$  & $27.7$ & $19.2$ \\ \hline
\multirow{2}{*}{7}
                             & $848 \times 480$   & $79.6$  & $50.4$  & $27.4$ & $19.2$ \\ \cline{2-6}
                             & $424 \times 240$   & $79.8$  & $50.0$  & $27.4$ & $19.3$ \\ \hline
\multirow{2}{*}{15}
                             & $848 \times 480$   & $200.0$ & $117.0$ & $59.6$ & $38.3$ \\ \cline{2-6}
                             & $424 \times 240$   & $200.7$  & $117.4$ & $60.1$ & $39.0$ \\ \hline
\end{tabular}
\end{minipage}
\caption{\label{tab:execution-time}Average ``execution time per frame'' (in milliseconds) with respect to the tested parameters}
\end{table}

\begin{table}[!htbp]
\begin{minipage}{.48\textwidth}
\centering
\begin{tabular}{|r|c|c c c c|}
\hline
\multicolumn{1}{|c|}{\bf{Win.}} & \diagbox{\bf{Image}}{\bf{Fps}}
                                             & 3       & 6       & 15      & 30      \\ \hline
\multirow{2}{*}{5}
                             & $848 \times 480$   & $0.58$ & $0.35$ & $0.18$ & $0.11$ \\ \cline{2-6}
                             & $424 \times 240$   & $0.59$ & $0.35$ & $0.18$ & $0.11$ \\ \hline
\multirow{2}{*}{7}
                             & $848 \times 480$   & $0.58$ & $0.34$ & $0.18$ & $0.11$ \\ \cline{2-6}
                             & $424 \times 240$   & $0.58$ & $0.35$ & $0.17$ & $0.11$ \\ \hline
\multirow{2}{*}{15}
                             & $848 \times 480$   & $0.61$ & $0.36$ & $0.18$ & $0.12$ \\ \cline{2-6}
                             & $424 \times 240$   & $0.62$ & $0.36$ & $0.19$ & $0.12$ \\ \hline
\end{tabular}
\end{minipage}
\caption{\label{tab:keyframe-per-frame}Average ``keyframe to frame ratio'' with respect to the tested parameters}
\end{table}

To determine suitable parameters for real-time execution, we calculate the speed
factor, which compares the image sampling time to the Ceiling-DSO execution
time. The speed factor, denoted as \(\kappa\), is defined as \(\kappa = \dfrac{1}{t_{F} \cdot
f}\), where \(t_F\) represents the per-frame execution time in seconds and \(f\) is
the frame rate in hertz (both values can obtained from table
\ref{tab:execution-time}). A speed factor greater than one (\(\kappa > 1\)) indicates that
the algorithm can process images faster than the frame rate, while a speed
factor less than one (\(\kappa < 1\)) implies that the algorithm cannot process images
at the frame rate speed, thereby failing to meet the timing constraints for
online processing.

\begin{figure}[htbp]
\centering
\includegraphics[width=0.6\textwidth]{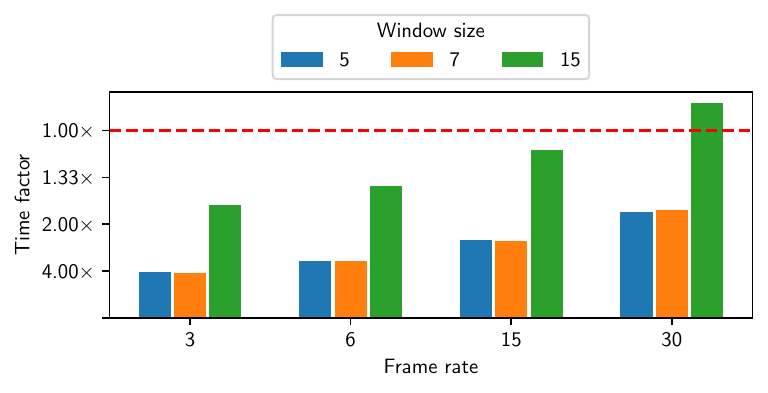}
\caption{\label{fig:speed-factor}Average \emph{speed factor}. The red dashed line delimits the real-time limit.}
\end{figure}

Figure \ref{fig:speed-factor} illustrates the speed factor for various combinations of
the tested parameters. As indicated in the table \ref{tab:execution-time}, there is no
apparent correlation between image size and execution time, so the bar plot in
the figure marginalizes the influence of image size. The results demonstrate
acceptable timing performance, except for the case of a high frame rate of 30fps
combined with a maximum window size of 15. From our tests, a window size of 7
appears to offer a good tradeoff between accuracy and runtime.

\section{Conclusions}
\label{sec:org09f8fde}
\label{sec:conclusion}

In this paper, we introduced Ceiling-DSO, an ceiling-vision odometry system
based on Direct Sparse Odometry (DSO) specifically designed for applications in
indoor mobile industrial robots. One of the key advantages of our system is its
suitability for dynamic environments commonly found in industrial settings.
Unlike other approaches, our system does not rely on assumptions about the
specific shape or content of the ceiling, making it a versatile solution. To
validate our approach, we conducted a comparison between the estimated
trajectories and the ground truth. Additionally, we conducted experiments to
analyze the impact of varying \emph{input image size}, \emph{input frame rate}, and
\emph{optimization window size} on the system's real-time capabilities, with the aim of
identifying an optimal parameter combination.

The experimental results indicated that altering the input image size did not
have a significant impact on the system's performance. However, changing the
input frame rate had a slight effect on the estimated trajectory and run time.
Based on our tests, a frame rate of 15fps offered a good balance between
accuracy and run time efficiency. Regarding the optimization window size,
varying it did not significantly affect the trajectory accuracy but had a
noticeable impact on run time. Our experiments demonstrated that using a maximum
window size of 7 provided satisfactory results in terms of both timing and
accuracy.

This work serves as a foundation for future research. Our future work will focus
on developing an online metric scale estimation approach based on sensor fusion.
Additionally, we aim to propose map management and loop-closing strategies to
offer a comprehensive SLAM solution for ceiling-vision applications.
Furthermore, we plan to make our curated dataset publicly available, enabling
other researchers to evaluate and validate their ceiling-vision algorithms using
real-world scenarios.

\vspace{6pt}

\authorcontributions{
Conceptualization, A.B., S.B. and E.S.; methodology, A.B. and E.S.; software, A.B.;
validation, A.B. and E.S.; formal analysis, A.B.;
resources, S.B. and F.G.; data curation, A.B.; writing---original draft preparation,
A.B.; writing---review and editing, E.S. and S.B.; visualization, A.B.; supervision,
S.B., E.S and F.G.; funding acquisition, S.B, E.S and F.G. All authors have
read and agreed to the published version of the manuscript.
}

\funding{
This research received no external funding.
}

\conflictsofinterest{
The authors declare no conflicts of interest.
}
\begin{adjustwidth}{-\extralength}{0cm}

\reftitle{References}

\bibliographystyle{unsrtnat}
\bibliography{library,/home/hacko/Zotero/library}

\begin{thebibliography}{999}

\bibitem[Bougouffa et~al.()Bougouffa, Seignez, Bouaziz, and
  Gardes]{bougouffa-evaluation-2022}
Bougouffa, A.; Seignez, E.; Bouaziz, S.; Gardes, F.
\newblock Evaluation of a Novel {{DSO-based Indoor Ceiling-Vision Odometry
  System}}.
\newblock In Proceedings of the 2022 17th {{International Conference}} on
  {{Control}}, {{Automation}}, {{Robotics}} and {{Vision}} ({{ICARCV}}), pp.
  47--53.
\newblock {\url{https://doi.org/10.1109/ICARCV57592.2022.10004272}}.

\bibitem[Qi et~al.()Qi, Yang, and Zhou]{qi-application-2015}
Qi, B.; Yang, Q.; Zhou, Y.Y.
\newblock Application of {{AGV}} in Intelligent Logistics System.
\newblock In Proceedings of the Fifth {{Asia International Symposium}} on
  {{Mechatronics}} ({{AISM}} 2015), pp. 1--5.
\newblock {\url{https://doi.org/10.1049/cp.2015.1527}}.

\bibitem[Borenstein()]{borenstein-omnimate-2000}
Borenstein, J.
\newblock The {{OmniMate}}: {{A}} Guidewire- and Beacon-Free {{AGV}} for Highly
  Reconfigurable Applications.
\newblock {\em 38},~1993--2010.
\newblock {\url{https://doi.org/10.1080/002075400188456}}.

\bibitem[Han et~al.()Han, Qian, Chung, Hou, Lee, Chen, Zhang, and
  Xu]{han-system-2013}
Han, L.; Qian, H.; Chung, W.K.; Hou, K.W.; Lee, K.H.; Chen, X.; Zhang, G.; Xu,
  Y.
\newblock System and Design of a Compact and Heavy-Payload {{AGV}} System for
  Flexible Production Line.
\newblock In Proceedings of the 2013 {{IEEE International Conference}} on
  {{Robotics}} and {{Biomimetics}} ({{ROBIO}}), pp. 2482--2488.
\newblock {\url{https://doi.org/10.1109/ROBIO.2013.6739844}}.

\bibitem[Siegwart et~al.()Siegwart, Nourbakhsh, and
  Scaramuzza]{siegwart-introduction-2011}
Siegwart, R.; Nourbakhsh, I.R.; Scaramuzza, D.
\newblock {\em Introduction to {{Autonomous Mobile Robots}}}, 2 ed.; The MIT
  Press.

\bibitem[Scaramuzza and Fraundorfer()]{scaramuzza-visual-2011}
Scaramuzza, D.; Fraundorfer, F.
\newblock Visual {{Odometry}} [{{Tutorial}}].
\newblock {\em 18},~80--92.
\newblock {\url{https://doi.org/10.1109/MRA.2011.943233}}.

\bibitem[Nister et~al.()Nister, Naroditsky, and Bergen]{nister-visual-2004}
Nister, D.; Naroditsky, O.; Bergen, J.
\newblock Visual Odometry.
\newblock In Proceedings of the Proceedings of the 2004 {{IEEE Computer Society
  Conference}} on {{Computer Vision}} and {{Pattern Recognition}}, 2004.
  {{CVPR}} 2004., Vol.~1, pp. I--I.
\newblock {\url{https://doi.org/10.1109/CVPR.2004.1315094}}.

\bibitem[Mur-Artal and Tard\'os()]{mur-artal-orbslam2-2017}
Mur-Artal, R.; Tard\'os, J.D.
\newblock {{ORB-SLAM2}}: {{An Open-Source SLAM System}} for {{Monocular}},
  {{Stereo}}, and {{RGB-D Cameras}}.
\newblock {\em 33},~1255--1262.
\newblock {\url{https://doi.org/10.1109/TRO.2017.2705103}}.

\bibitem[Awang~Salleh and Seignez()]{awangsalleh-swift-2018}
Awang~Salleh, D.N.S.D.; Seignez, E.
\newblock Swift {{Path Planning}}: {{Vehicle Localization}} by {{Visual
  Odometry Trajectory Tracking}} and {{Mapping}}.
\newblock {\em 06},~221--230.
\newblock {\url{https://doi.org/10.1142/S2301385018500085}}.

\bibitem[Davison et~al.()Davison, Reid, Molton, and
  Stasse]{davison-monoslam-2007}
Davison, A.J.; Reid, I.D.; Molton, N.D.; Stasse, O.
\newblock {{MonoSLAM}}: {{Real-Time Single Camera SLAM}}.
\newblock {\em 29},~1052--1067.
\newblock {\url{https://doi.org/10.1109/TPAMI.2007.1049}}.

\bibitem[Awang~Salleh and Seignez()]{awangsalleh-longitudinal-2019}
Awang~Salleh, D.N.S.D.; Seignez, E.
\newblock Longitudinal Error Improvement by Visual Odometry Trajectory Trail
  and Road Segment Matching.
\newblock {\em 13},~313--322.
\newblock {\url{https://doi.org/10.1049/iet-its.2018.5272}}.

\bibitem[Engel et~al.()Engel, Koltun, and Cremers]{engel-direct-2018}
Engel, J.; Koltun, V.; Cremers, D.
\newblock Direct {{Sparse Odometry}}.
\newblock {\em 40},~611--625.
\newblock {\url{https://doi.org/10.1109/TPAMI.2017.2658577}}.

\bibitem[Forster et~al.()Forster, Pizzoli, and Scaramuzza]{forster-svo-2014}
Forster, C.; Pizzoli, M.; Scaramuzza, D.
\newblock {{SVO}}: {{Fast}} Semi-Direct Monocular Visual Odometry.
\newblock In Proceedings of the 2014 {{IEEE International Conference}} on
  {{Robotics}} and {{Automation}} ({{ICRA}}), pp. 15--22.
\newblock {\url{https://doi.org/10.1109/ICRA.2014.6906584}}.

\bibitem[Geiger et~al.()Geiger, Ziegler, and Stiller]{geiger-stereoscan-2011}
Geiger, A.; Ziegler, J.; Stiller, C.
\newblock {{StereoScan}}: {{Dense}} 3d Reconstruction in Real-Time.
\newblock In Proceedings of the 2011 {{IEEE Intelligent Vehicles Symposium}}
  ({{IV}}), pp. 963--968.
\newblock {\url{https://doi.org/10.1109/IVS.2011.5940405}}.

\bibitem[Engel et~al.()Engel, Sch\"ops, and Cremers]{engel-lsdslam-2014}
Engel, J.; Sch\"ops, T.; Cremers, D.
\newblock {{LSD-SLAM}}: {{Large-Scale Direct Monocular SLAM}}.
\newblock In Proceedings of the Computer {{Vision}} -- {{ECCV}} 2014; Fleet,
  D.; Pajdla, T.; Schiele, B.; Tuytelaars, T., Eds. Springer International
  Publishing, Lecture {{Notes}} in {{Computer Science}}, pp. 834--849.
\newblock {\url{https://doi.org/10.1007/978-3-319-10605-2_54}}.

\bibitem[Sheng et~al.()Sheng, Pan, Gao, Tan, and Zhao]{sheng-dynamicdso-2020}
Sheng, C.; Pan, S.; Gao, W.; Tan, Y.; Zhao, T.
\newblock Dynamic-{{DSO}}: {{Direct Sparse Odometry Using Objects Semantic
  Information}} for {{Dynamic Environments}}.
\newblock {\em 10},~1467.
\newblock {\url{https://doi.org/10.3390/app10041467}}.

\bibitem[Kim and Kim()]{kim-effective-2016}
Kim, D.H.; Kim, J.H.
\newblock Effective {{Background Model-Based RGB-D Dense Visual Odometry}} in a
  {{Dynamic Environment}}.
\newblock {\em 32},~1565--1573.
\newblock {\url{https://doi.org/10.1109/TRO.2016.2609395}}.

\bibitem[WooYeon and Kyoung()]{wooyeon-cvslam-2005}
WooYeon, J.; Kyoung, M.L.
\newblock {{CV-SLAM}}: A New Ceiling Vision-Based {{SLAM}} Technique.
\newblock In Proceedings of the 2005 {{IEEE}}/{{RSJ International Conference}}
  on {{Intelligent Robots}} and {{Systems}}, pp. 3195--3200.
\newblock {\url{https://doi.org/10.1109/IROS.2005.1545443}}.

\bibitem[Kim et~al.()Kim, Choi, Lee, and Kim]{kim-new-2013}
Kim, D.Y.; Choi, H.; Lee, H.; Kim, E.
\newblock A New {{cvSLAM}} Exploiting a Partially Known Landmark Association.
\newblock {\em 27},~1073--1086.
\newblock {\url{https://doi.org/10.1080/01691864.2013.805470}}.

\bibitem[Hwang and Song()]{hwang-monocular-2011}
Hwang, S.; Song, J.
\newblock Monocular {{Vision-Based SLAM}} in {{Indoor Environment Using
  Corner}}, {{Lamp}}, and {{Door Features From Upward-Looking Camera}}.
\newblock {\em 58},~4804--4812.
\newblock {\url{https://doi.org/10.1109/TIE.2011.2109333}}.

\bibitem[Choi et~al.({\natexlab{a}})Choi, Kim, Hwang, Park, and
  Kim]{choi-efficient-2012}
Choi, H.; Kim, D.Y.; Hwang, J.P.; Park, C.W.; Kim, E.
\newblock Efficient {{Simultaneous Localization}} and {{Mapping Based}} on
  {{Ceiling-View}}: {{Ceiling Boundary Feature Map Approach}}.
\newblock {\em 26},~653--671.
\newblock {\url{https://doi.org/10.1163/156855311X617542}}.

\bibitem[Choi et~al.({\natexlab{b}})Choi, Kim, and Kim]{choi-efficient-2014}
Choi, H.; Kim, R.; Kim, E.
\newblock An {{Efficient Ceiling-view SLAM Using Relational Constraints Between
  Landmarks}}:.
\newblock {\url{https://doi.org/10.5772/57225}}.

\bibitem[Ribacki et~al.()Ribacki, Jorge, Mantelli, Maffei, and
  Prestes]{ribacki-visionbased-2018}
Ribacki, A.; Jorge, V.A.M.; Mantelli, M.; Maffei, R.; Prestes, E.
\newblock Vision-{{Based Global Localization Using Ceiling Space Density}}.
\newblock In Proceedings of the 2018 {{IEEE International Conference}} on
  {{Robotics}} and {{Automation}} ({{ICRA}}), pp. 3502--3507.
\newblock {\url{https://doi.org/10.1109/ICRA.2018.8460515}}.

\bibitem[Li et~al.()Li, Zhu, Yu, and Wang]{li-improved-2018}
Li, Y.; Zhu, S.; Yu, Y.; Wang, Z.
\newblock An Improved Graph-Based Visual Localization System for Indoor Mobile
  Robot Using Newly Designed Markers.
\newblock {\em 15},~1729881418769191.
\newblock {\url{https://doi.org/10.1177/1729881418769191}}.

\bibitem[Schubert et~al.()Schubert, Demmel, Usenko, St\"uckler, and
  Cremers]{schubert-direct-2018}
Schubert, D.; Demmel, N.; Usenko, V.; St\"uckler, J.; Cremers, D.
\newblock Direct {{Sparse Odometry}} with {{Rolling Shutter}}.
\newblock In Proceedings of the Computer {{Vision}} -- {{ECCV}} 2018; Ferrari,
  V.; Hebert, M.; Sminchisescu, C.; Weiss, Y., Eds. Springer International
  Publishing, Lecture {{Notes}} in {{Computer Science}}, pp. 699--714.
\newblock {\url{https://doi.org/10.1007/978-3-030-01237-3_42}}.

\bibitem[Engel et~al.()Engel, Usenko, and Cremers]{engel-photometrically-2016}
Engel, J.; Usenko, V.; Cremers, D.
\newblock A {{Photometrically Calibrated Benchmark For Monocular Visual
  Odometry}},  \href{http://arxiv.org/abs/1607.02555}{{\normalfont
  [1607.02555]}}.
\newblock {\url{https://doi.org/10.48550/arXiv.1607.02555}}.

\bibitem[Huber()]{huber-robust-1964}
Huber, P.J.
\newblock Robust {{Estimation}} of a {{Location Parameter}}.
\newblock {\em 35},~73--101.
\newblock {\url{https://doi.org/10.1214/aoms/1177703732}}.

\bibitem[Sol\`a et~al.()Sol\`a, Deray, and Atchuthan]{sola-micro-2020}
Sol\`a, J.; Deray, J.; Atchuthan, D.
\newblock A Micro {{Lie}} Theory for State Estimation in Robotics,
  \href{http://arxiv.org/abs/1812.01537}{{\normalfont [1812.01537]}}.
\newblock {\url{https://doi.org/10.48550/arXiv.1812.01537}}.

\bibitem[Bougouffa et~al.()Bougouffa, Seignez, Bouaziz, and
  Gardes]{bougouffa-smarttrolley-2020}
Bougouffa, A.; Seignez, E.; Bouaziz, S.; Gardes, F.
\newblock {{SmartTrolley}}: {{An Experimental Mobile Platform}} for {{Indoor
  Localization}} in {{Warehouses}}.
\newblock In Proceedings of the 2020 3rd {{International Conference}} on
  {{Robotics}}, {{Control}} and {{Automation Engineering}} ({{RCAE}}), pp.
  108--115.
\newblock {\url{https://doi.org/10.1109/RCAE51546.2020.9294484}}.

\bibitem[Pedrosa et~al.()Pedrosa, Pereira, and Lau]{pedrosa-fast-2020}
Pedrosa, E.; Pereira, A.; Lau, N.
\newblock Fast {{Grid SLAM Based}} on {{Particle Filter}} with {{Scan
  Matching}} and {{Multithreading}}.
\newblock In Proceedings of the 2020 {{IEEE International Conference}} on
  {{Autonomous Robot Systems}} and {{Competitions}} ({{ICARSC}}). IEEE, pp.
  194--199.
\newblock {\url{https://doi.org/10.1109/ICARSC49921.2020.9096191}}.

\bibitem[Zhang and Scaramuzza()]{zhang-tutorial-2018}
Zhang, Z.; Scaramuzza, D.
\newblock A {{Tutorial}} on {{Quantitative Trajectory Evaluation}} for
  {{Visual}}(-{{Inertial}}) {{Odometry}}.
\newblock In Proceedings of the 2018 {{IEEE}}/{{RSJ International Conference}}
  on {{Intelligent Robots}} and {{Systems}} ({{IROS}}), pp. 7244--7251.
\newblock {\url{https://doi.org/10.1109/IROS.2018.8593941}}.

\end{thebibliography}

\PublishersNote{}
\end{adjustwidth}
\end{document}